# A Learning-Based Visual Saliency Prediction Model for Stereoscopic 3D Video (LBVS-3D)

Amin Banitalebi-Dehkordi, Mahsa T. Pourazad, and Panos Nasiopoulos

**Abstract** Saliency prediction models provide a probabilistic map of relative likelihood of an image or video region to attract the attention of the human visual system. Over the past decade, many computational saliency prediction models have been proposed for 2D images and videos. Considering that the human visual system has evolved in a natural 3D environment, it is only natural to want to design visual attention models for 3D content. Existing monocular saliency models are not able to accurately predict the attentive regions when applied to 3D image/video content, as they do not incorporate depth information. This paper explores stereoscopic video saliency prediction by exploiting both low-level attributes such as brightness, color, texture, orientation, motion, and depth, as well as high-level cues such as face, person, vehicle, animal, text, and horizon. Our model starts with a rough segmentation and quantifies several intuitive observations such as the effects of visual discomfort level, depth abruptness, motion acceleration, elements of surprise, size and compactness of the salient regions, and emphasizing only a few salient objects in a scene. A new fovea-based model of spatial distance between the image regions is adopted for considering local and global feature calculations. To efficiently fuse the conspicuity maps generated by our method to one single saliency map that is highly correlated with the eye-fixation data, a random forest based algorithm is utilized. The performance of the proposed saliency model is evaluated against the results of an eye-tracking experiment, which involved 24 subjects and an in-house database of 61 captured stereoscopic videos. Our stereo video database as well as the eye-tracking data are publicly available along with this paper. Experiment results show that the proposed saliency prediction method achieves competitive performance compared to the state-of-the-art approaches.[1]

# 1 Introduction

When watching natural scenes, an overwhelming amount of information is delivered to the human eye, with the optic nerve receiving an estimated $10^8$ bits of information per second [1]. In order for the Human Visual System (HVS) to process this volume of visual data, it separates the data into pre-attentive and attentive levels [2]. The former is responsible for identifying the regions worth of attention, while the later involves in-depth processing of limited portions of the visual information [2].

In computer vision, there is a strong interest in designing models inspired by HVS that narrow down a large amount of visual data to smaller amount of more visually important data. Generally, eye-tracking experiments are used to help us understand what catches human attention in a scene. However, eye-tracking devices are not a viable option in many automated applications. Instead, Visual Attention Models (VAMs) have been developed to mimic the layered perception mechanism of the human visual system by automatically detecting Regions Of Interest (ROIs) in a scene.

Psychological findings suggest that, in the pre-attentive stage, the visual information of a scene is represented by several retinotopic maps, each of them illustrating one visual attribute [3]. These attributes along with higher-level scene-dependent information are then analyzed by the visual cortex. Motivated by this fact, visual attention models also predict the locations of salient regions using three different approaches: bottom-up, top-down, and integration of the two. Bottom-up saliency detection models adopt rapid low-level visual attributes such as brightness, color, motion, and texture to generate a stimulus driven saliency map. Top-down approaches, however, utilize high-level context-dependent information such as humans, faces, animals, cars, and text for saliency detection in specific tasks. Integrated methods utilize bottom-up and top-down attributes for saliency detection [4].

There has been a great deal of research done in the field of 2D images and video saliency analysis that resulted in developing many successful visual attention models for 2D content [5-16], [82-85]. However, two-dimensional VAMs are usually not accurate enough in predicting the salient regions in 3D content, as they do not incorporate the depth information [17-20]. One reason is that depth perception changes the impact of the 2D visual saliency attributes (e.g., brightness, color, texture, motion). Also, there are several other visual attributes such as depth range, display size, the technology used in 3D display (i.e., active or passive glasses, glasses-free auto-stereoscopic displays, etc.), naturalness [21], and visual comfort [22] that solely affect 3D attention while they don't have any impact on 2D visual attention [23-24]. As a result, in order to truly mimic the visual attention of the HVS, we need to use 3D-exclusive saliency prediction mechanisms. The points made above and the rapid expansion of 3D image and video technologies emphasize the necessity to either extend the current 2D saliency detection mechanisms to 3D data, or develop novel 3D-specific saliency prediction methods.

[1] This work was partly supported by Natural Sciences and Engineering Research Council of Canada (NSERC) under Grant STPGP 447339-13 and Institute for Computing Information and Cognitive Systems (ICICS) at UBC.
  M. T. Pourazad is with ICICS at the University of British Columbia (UBC) and TELUS Communications Inc., Canada (pourazad@icics.ubc.ca). The other authors are with the Electrical Engineering Department and ICICS at UBC, Vancouver, BC, Canada ({dehkordi, panos}@ece.ubc.ca).

The existing literature for 3D saliency prediction offers two main groups of solutions. The first group of solutions (earliest attempts for 3D saliency prediction) directly uses the depth map (or disparity map) as a weighting factor in conjunction with an existing 2D saliency detection model. In other words, a 2D saliency map is created first, then each pixel (or region) in the resulting map is assigned a weight according to its depth (disparity) value. Maki et al. [25], Zhang et al. [26], and Chamaret et al. [27] used this approach to design their computational model for 3D saliency. They mainly design their methods based on the idea that generally the objects that are closer to the observer are considered to be more salient. Although they observed qualitative improvements compared to 2D saliency mechanisms, they didn't provide a quantitative evaluation of the proposed methods. Moreover, objects closer to viewers are not necessarily more salient. Fig. 1 shows an example where although the "ground" is the closest object to viewers, visual attention is directed to other parts of the scene.

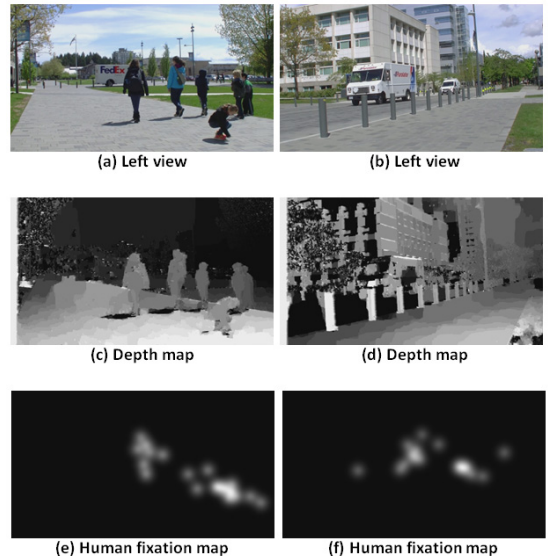

**Fig. 1.** Closer objects are not necessarily salient. "Ground" is not considered salient in these examples.

The second group of solutions for 3D visual attention prediction utilizes the depth information of a scene to create a depth saliency map. The depth saliency map is usually combined with the 2D conspicuity maps (using the existing visual attention models) to construct a computational model of 3D visual saliency. Using this approach, Ouerhani and Hugli [28] proposed an attention model that takes into account the depth gradient features as well as the surface curvature. They performed qualitative assessment, but no quantitative assessment against eye-tracking data was done. Lang et al. [29] proposed a depth saliency map in which they evaluated the statistical probability of saliency ratio at different depth ranges using a training database. To validate their method, they also integrated the resulting depth saliency map to some other 2D models by summation or element-wise multiplication. Wang et al. [18] incorporated a Bayesian approach of depth saliency map generation and combined their map with some existing 2D models through averaging. Fang et al. [17] proposed a computational model of saliency for stereoscopic images by taking into account four different attributes: brightness, color, texture, and depth. They partitioned each image into patches and considered DC and AC coefficients of the DCT transform of each patch as its corresponding features. They generated several feature maps and linearly combined them with an emphasis on the compactness property of feature maps. Unlike the above methods that were designed for stereo images, Kim et al.'s work [20] was among a few saliency prediction models, which were proposed for stereoscopic videos. They adopted a scene type classification mechanism and incorporated several saliency attributes as well as concepts like saliency compactness, depth discontinuities, and visual discomfort. The generated feature maps were combined through summation or element-wise multiplication. It is common practice for saliency prediction methods to calculate various feature maps and then average them into one final map. However, it is not exactly known how the human brain fuses the different visual attributes. Examining the importance of each of the features and determining how to properly fuse them to closely imitate the human visual system remains a challenge. Another challenge in stereo video saliency prediction is that the accuracy of the existing visual attention models are still not very high, especially when compared to the accuracy of 2D visual attention models for 2D video [57]. Moreover, most of the existing methods do not provide enough flexibility to be tailored for specific types of video scene content (e.g. large motion, natural or synthetic, appearance of high-level attributes like humans, and etc.).

This paper investigates the computational modeling of visual attention of stereoscopic video by proposing an integrated saliency prediction method. The proposed approach utilizes both low-level attributes such as brightness, color, texture, orientation, motion, and depth as well as high-level context-dependent cues such as face, person, vehicle, animal, text, and horizon. Our model starts with a rough segmentation and quantifies several intuitive observations such as the effects of visual discomfort level, depth abruptness, motion acceleration, elements of surprise and size, compactness, and sparsity of the salient regions. To calculate local and global features describing these observations, a new fovea-based model of spatial distance between the image regions is used. Then, a random forest based algorithm is utilized to learn a model of stereoscopic video saliency so that the various conspicuity maps generated by our method are efficiently fused into one single saliency map, which delivers high correlation with the eye-fixation data. The performance of the proposed model is evaluated against the results of a large-scale eye-tracking experiment, which involves 24 subjects and an in-house database of 61 captured stereoscopic videos. This database is made publicly available.

The main contributions of this paper are as follows: 1) Using several (some existing and some proposed) low-level and high-level indicative saliency features that have potential in predicting salient regions in a 3D scene. 2) Taking into account the effect of biological and intuitive observations in 3D saliency prediction, by modeling natural elements of human visual system into saliency features. 3) Combining the candidate saliency features in a learning-based framework to generate an overall saliency map per frame which has high correlation with the eye-tracking data. 4) Using the results of eye-tracking experiments to analyze: the contribution of each feature, benchmarking the performance of various 3D VAMs, and evaluate the statistical difference between the 3D VAMs.

We propose a flexible framework to learn from any desired saliency feature, to predict saliency for unseen stereo video data. We also propose a suggested set of low-level attributes such as brightness, color, texture, orientation, motion, and depth as well as high-level context-dependent cues such as face, person, vehicle, animal, text, and horizon. These are suggested set of features, however, any other feature can be added or removed.

The rest of this paper is organized as follows: Section 2 explains the proposed saliency prediction method, Section 3 elaborates on the database creation, and subjective tests, results and discussions are provided in Section 4, and Section 5 concludes the paper.

## 2 Proposed saliency prediction method

Our proposed visual attention model takes into account various low-level saliency attributes as well as high-level context-dependent cues. In addition, several intuitive observations are quantified and considered in the design of our VAM. Once the feature maps are extracted, a random-forest-based algorithm is used to train a model of saliency prediction. Note that the extracted feature maps are "indicative" features that are likely to reflect the salient regions. The proposed learning-based fusion model combines the extracted features to generate the overall saliency maps. In Section 4, we provide the importance of each individual feature along with the additional computational complexity it causes. The flowchart of our method is illustrated in Fig. 2 and the following sections elaborate on details of our model.

### 2.1 Bottom-up saliency features

The proposed model includes luminance, color, texture, motion, and depth as low-level saliency features. Each of them is explain in the following sub-sections. Note that since the position of objects is slightly different between the left and right views, with the exception of depth features, the rest of the features are extracted from the view of the video for which the depth map is available. In our experiments, for each video both the left and right views are initially available. We calculate the left-to-right disparity (which corresponds to disparity map of the right view) using the Depth Estimation Reference Software (DERS) [30]. Therefore, the right view is used for computing the 2D saliency attributes. The motivation behind the selection of the right view is that humans are mostly right-eye dominant (approximately 70%) [31].

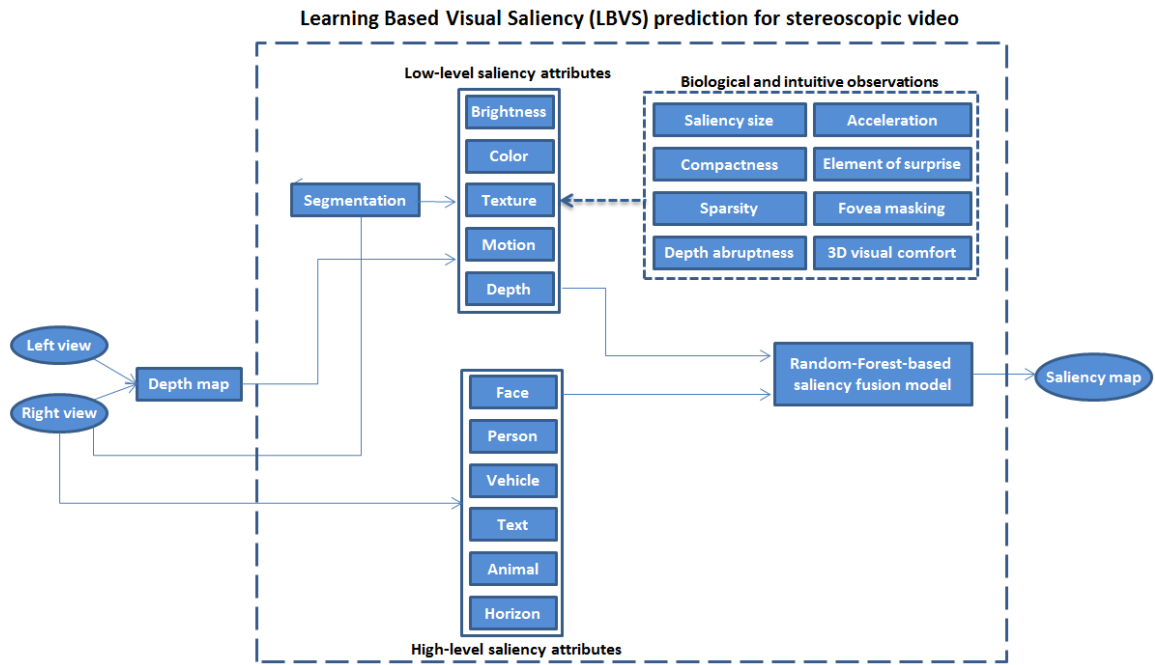

**Fig. 2.** Proposed computational model for visual saliency prediction for stereoscopic video.

*1) Segmentation*

During the pre-attentive stage of HVS, the visual information of each scene is partitioned to different regions. Computational resources are then allocated to each region based on its relative importance (when we look at a scene, we visualize it as a set of patches, not a grid of pixels). Similarly, we perform a rough segmentation on the right view picture and assign bottom-up saliency attributes to each segment separately, by averaging the specified pixel-wise saliency values over each segment. In our implementation, we use the Edge Detection and Image Segmentation (EDISON) System proposed by Comanicu et al. [32]. Note that similar to computational resource allocation performed by the HVS in the pre-attentive vision, we use the segmentation only for generating the saliency features. For combining the generated features and producing an overall saliency map, we use the proposed random forest based feature fusion model.

*2) Brightness map*

Studies have shown that human attention is directed towards areas with higher brightness variations in a scene [5,7]. In our model, we include a map of local brightness variances. To this end, each frame of the right view is transformed to the YUV color space first. Then, local variances are calculated in a circular neighborhood around each pixel using the following formula:

$$\sigma_x^2 = \frac{1}{size_{\{NHOOD\}} - 1} \sum_{i \in NHOOD} (Y_x - Y_i)^2 \quad (1)$$

where $Y_x$ and $Y_i$ represent the brightness intensity values of the center pixel and surrounding pixels respectively, *NHOOD* is a circular neighborhood, $x$ denotes the current pixel, and $i$ is pixel index over the outer circular area. The choice of size and shape of the outer area used for variance calculation is based on the fovea photoreceptor concentration, which is explained in Section 2.1.7. The variance map for each frame is normalized to [0-1]. The resulting brightness variance map contains local brightness variances. However, attention is directed towards certain areas of an image based on both local and global scene properties. Therefore, we adjust the value of the brightness variance map for each pixel as follows: For each pixel, the weighted average difference between its brightness variance and the brightness variance of the surrounding pixels is calculated, resulting in a brightness variance contrast map. This process is known as "center-surround operation", and the Differences-of-Gaussians (DoG) are common approaches for computing differences [5]. Here, instead of the Gaussians, we use a circular mask, which is designed based on fovea photoreceptor concentration (See II.*A*.7). Hereafter, we refer to this mask as "fovea mask". The fovea mask assigns different weights to different pixels based on their distance from the pixel located at the center of the mask. The center-surround operation for globalizing the variance map is performed as follows:

$$S_{Brightness}^{K} = \frac{1}{n_K} \sum_{i \in R_K} \sum_{\substack{j \in fovea\ mask \\ j \neq i}} \left( |Var_{diff}(i,j)| \times Fovea_{mask}(j) \right) \quad (2)$$

where $S^{K}_{Brightness}$ is the saliency value assigned to the $K^{th}$ region ($R_K$), $n_K$ is the number of pixels in the $K^{th}$ region, $i$ denotes the $i^{th}$ pixel in the current segment, $j$ is the $j^{th}$ pixel in the fovea mask centered at $i$, and $Var_{diff}$ is the difference between $i^{th}$ and $j^{th}$ pixels in the brightness variance map (obtained from (1)). The center-surround operand results in a brightness variance contrast map in which every segment (or roughly every object) is assigned with one saliency probability.

It is worth noting that while the proposed brightness variance contrast map captures the second order variations in brightness (contrast in the variance of brightness), we also follow what is considered a common practice in literature by adding a brightness contrast map that captures the first order brightness variations in a scene. Brightness contrast for each region is calculated similar to the brightness variance contrast map mentioned above. The only difference is the usage of the brightness contrast instead of the brightness variance contrast.

*3) Color maps*

Color is one of the most important channels among human senses, accounting for 80% of the visual experience [33]. For the proposed model, the following feature maps are extracted from the color information of each scene:

<u>Color histogram map</u>: Naturally, humans tend to look at the objects with colors that stand out in a scene, i.e., rare colors tend to attract human attention. To account for the effect of color rarity, we compute a color histogram for each picture, based on the occurrence probability of each of the three color channels in the RGB space (each channel represented by 8 bits). Suppose $p(R=r, G=g, B=b)=P$. We define the values of $e^{\frac{-P}{\overline{P}}}$ as a saliency map related to color rarity and call it "histogram saliency map" ($\overline{P}$ is the average of $P$). The exponential function is particularly chosen to project the probability values in the interval [0-1] and exposes a local maximum where a color is rare.

<u>Color variance contrast maps</u>: HVS is highly sensitive to color contrast. Similar to the brightness variance contrast map generated before (see *II.A.2*), two new color saliency maps are created for a* and b* color components in the CIE L*a*b* domain. This color space is particularly chosen because of its uniform chromaticity properties [34].

<u>Warmth and saturation color maps</u>: Experiments showed that warm and saturated colors are generally salient to HVS [35-36].

Warm colors dominate their surroundings regardless of the existence of color contrast in the background [36]. Highly bright and saturated colors are salient regardless of their associated hue value [36]. The reason is partially due to the eye sensitivity to these types of colors [34], and partially due to human natural instincts, which interpret warm and saturated colors (e.g., saturated red) as a potential threat. To account for the effect of warm colors, the color temperature of the right view picture is calculated first. In general, warm colors correspond to low temperatures and vice versa. The color warmth map is defined as the inverse of the color temperature map. To account for saliency based on saturation of colors, we follow the saturation formula of Lübbe [37]:

$$Saturation\ map = \frac{C^*_{ab}}{\sqrt{C^{*2}_{ab} + L^{*2}}} \qquad (3)$$

where $L^*$ is the lightness (brightness) and $C^*_{ab}$ denotes the chroma of the color calculated as:

$$C^*_{ab} = \sqrt{a^{*2} + b^{*2}} \qquad (4)$$

HVS color sensitivity map: Generally, human eyes have different perception sensitivity at different light wavelengths. We use the CIE 1978 spectral sensitivity function values [34] at different wavelengths to locate the image regions for which eyes are more sensitive to the light. To this end, the dominant wavelength for a table of spectral colors (monochromatic colors) is computed first [38]. We assume that the image colors are monochrome (note that conversion from RGB values to light wavelength is not possible for non-monochrome colors, as each color can be represented by many combinations of R, G, and B values at different wavelengths). Then, for each pixel of the right view, the closest spectral color and thus its associated dominant wavelength are selected from the available look-up table. The eye sensitivity at each wavelength is depicted as a map of sensitivity probability.

Empirical color saliency map: Several subjective studies have been carried out to test the visual attention saliency of colors using eye-tracking information [35,39]. In these studies, shapes of different colors are shown to the viewers and based on the eye fixation statistics conclusions are made on the saliency of various colors. Gelasca et al. [39] sorted 12 different colors based on their received visual attention. We use the results of their experiment to build a look-up table for these colors. Then, for each pixel within the right view picture, a saliency probability is assigned based on the closest numerical distance of the RGB values of that pixel and the table entries (mean squared error of the R, G, and B values).

It is worth noting that we use the RGB color space for the color histogram map and empirical color saliency map, as for the former case we don't use the color differences directly, and for the latter case we use the empirical color saliency probabilities from 12 distinctly different colors. Our experiments verified the efficiency of our choice of color space for these features.

*4) Texture map*

Texture and orientation of picture elements are among the most important saliency attributes [4]. To generate a texture saliency map, we utilize the Gabor filters (which are widely used for texture extraction) to create a Gabor energy map at 4 different scales and 8 orientations. The *L2* norm of the Gabor coefficients results in a Gabor energy map. Since image texture is perceived locally at each instance of time, we apply our fovea mask to the Gabor energy map to generate a texture map that contains the edges and texture structure of the image. However, not every edge or image structure is salient. To emphasize salient edges and de-emphasize non-salient texture, we create an edginess map and multiply (element-wise) it by the current texture map. Edginess per unit area is defined by:

$$Edginess_K = \frac{1}{n_K} \sum_{R_K} Edgemap_K \qquad (5)$$

where $R_K$ denotes the region $K$, $n_K$ is number of pixels, and *Edgemap* contains the edges for the $K^{th}$ region. Note that edge maps are available as a part of the segmentation algorithm explained previously. Our choice of *Edgemap* ensures that areas with dense edges are assigned with higher saliency probabilities compared to areas with sparse edges.

*5) Motion maps*

Due to humans' biological instincts, moving objects always attract human attention. In 3D, there exist three different motion directions: horizontal (*dx*), vertical (*dy*), and perpendicular to the screen (*dz*). Moreover, since it is not clear which one of these directions or which combination of them implies higher impact on the visual saliency, we generate one motion vector for each direction, keep them as separate motion maps, and examine the importance of each attribute using our random forest learning algorithm (Section 2.3).

To extract horizontal and vertical motion maps (*Dx* and *Dy*) for right view frames, we incorporate the correlation flow algorithm by Drulea and Nedevschi [40], as this method has shown promising performance on various datasets and is publicly available. Motion along the *Z* direction particularly exists for 3D video and does not appear in the 2D case. To extract the motion vector along the *Z* direction, we utilize the available depth information (more details on the availability of depth data is presented in Section 2.1.6) as follows:

$$Dz = \text{Ref}_{Depth}(i_x, i_y) - \text{Current}_{Depth}(i_x + dx, i_y + dy) \quad (6)$$

where $Ref_{Depth}(i_x,i_y)$ is the depth value of the $i^{th}$ pixel of a reference frame (or previous frame) with horizontal and vertical location of $x$ and $y$, and $Current_{Depth}(i_x+dx,i_y+dy)$ is the depth value of the $i^{th}$ pixel in the current frame with horizontal and vertical coordinates of $i_x+dx$ and $i_y+dy$. Note that $dx$ and $dy$ are calculated using the optical flow algorithm of [40] and $(i_x+dx,i_y+dy)$ in the current frame is the approximate location of $(i_x,i_y)$ in the previous frame.

When training a regression algorithm (in machine learning in general), it is a common practice to normalize the feature values. This generally helps the training by avoiding bias towards a particular feature subset. Two most common ways of achieving this are: 1) normalization by scaling (e.g. scale to 0-1), and 2) standardization (use mean and standard deviation values of the features to form a standard Gaussian distribution). We choose the former approach to be consistent with other feature maps. To this end, once the $Dx$, $Dy$, and $Dz$ maps are generated, they are normalized to the interval of [0-1] (on a frame-by-frame basis). Note that in our experiments we found out that normalization slightly improves the regression accuracy. Also, note that we did not examine the standardization approach, or any other alternative method to relate $Dx$ and $Dy$ to $Dz$.

Next, for each segment of the right view, the average motion value is assigned to that segment under the assumption that object motions are homogeneous. Note that while the classical averaging operator is used here, median, mode, minimum, or any other meaningful operator can be also used. The following motion maps are used in our method:

Velocity: Velocity in different directions is defined as:

$$V_x = (fr-1)dx \;, \; V_y = (fr-1)dy \;, \; V_z = (fr-1)dz \quad (7)$$

where $fr$ is the frame rate of the stereoscopic video. Note that frame rate is of particular importance in 3D as motion highly affects the 3D video Quality of Experience (QoE) [41-42]. The velocity vector magnitude is evaluated as:

$$V = \sqrt{V_x^2 + V_y^2 + V_z^2} \quad (8)$$

Velocity with emphasize on $V_z$: Due to humans' survival instincts, objects that are on a collision path towards them are treated as a possible threat. Therefore, attention is directed towards them [43]. Inspired by this fact, we modify the velocity vector, emphasizing the velocity in the Z direction. Thus, $Dz$ is re-defined as:

$$Dz_1 = e^{dz} - 1 \quad (9)$$

Using the new $Dz$ value, the velocity vector in Z direction and then the velocity vector for each segment are re-calculated. The resulting map is referred to as $V_{z-emphasized}$.

Acceleration: Objects with relatively high acceleration are generally considered to be salient. We add a map of relative acceleration to our set of motion saliency maps by computing the acceleration using the following formula:

$$A = \frac{\Delta V}{\Delta t} = (fr-1)[V_{Current \atop frame} - V_{Reference \atop frame}] \quad (10)$$

Element of surprise: One of the differences between saliency prediction in images and videos is the possible introduction of an element of surprise in video, which turns the attention towards itself. To account for the effect of unusual motion, we emphasize on the saliency of segments with small motion vector occurrence probabilities. To this end, we define a motion histogram map in the range of [0-1] as:

$$M_{probability} = e^{\frac{-p(mv)}{\bar{p}}} \quad (11)$$

and use it as a feature for saliency prediction. In the above equation, $mv$ represents the motion vector $(dx, dy, dz)$, and $p(mv)$ is the joint probability density function of uniformly quantized 3D motion directions in the current stereo frame.

*6) Depth saliency map*

As mentioned previously, in our method we extract the left-to-right disparity map using the DERS [30] software. However, the same disparity values in a disparity map could correspond to different perceived depths depending on the viewing conditions. Therefore, we use the disparity map to generate a depth map for the right view picture. Fig. 3 illustrates two similar triangles when an observer is watching a 3D video. Writing down the trigonometry equations gives:

$$\begin{cases} \dfrac{disparity\,(W/R_W)}{L_{eyes}} = \dfrac{x}{D} \\ x = Z_{observer} - D \end{cases} \Rightarrow Depth = \dfrac{Z_{observer}}{1 + \dfrac{disparity \times W}{L_{eyes} \times R_W}} \quad (12)$$

where $Z_{observer}$ represents the distance of the viewer's eyes to 3D screen (183 *cm* in our experiments), $L_{eyes}$ is the inter-ocular distance between the two eyes (on average 6.3 *cm* for humans), and $W$ and $R_W$ are the horizontal width (in *cm*) and resolution (in pixels) of the display screen, respectively. Note that this method of disparity to depth conversion results in read-world depth values (in *cm*) and has been similarly used by other colleagues [17-18]. Also, note that our saliency prediction mechanism requires a disparity which does not necessarily have to be generated using DERS. Any disparity detection algorithm can be used for disparity map generation.

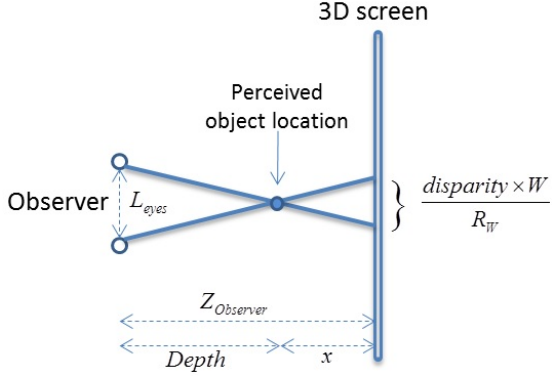 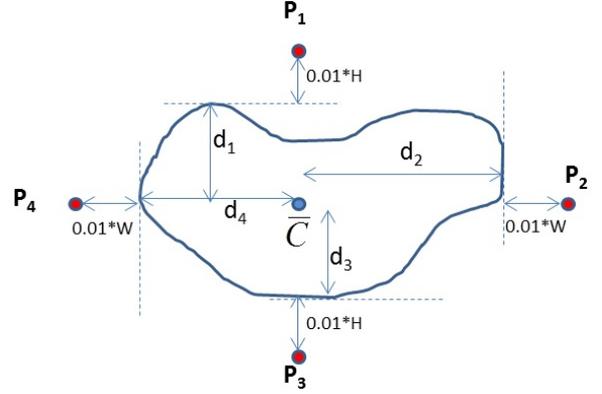

**Fig. 3.** Disparity to depth conversion scheme viewed from above.

**Fig. 4.** Creating a depth abruptness mask: An example segment with four selected points around it.

A depth map can be incorporated to create a depth saliency map in which closer objects are assigned higher saliency values. To create such a map, the inverse of depth values is linearly mapped to [0-1]. The resulting map, however, does not demonstrate the saliency perfectly since not every close object is salient (See Fig. 1 for a counter example). An object that stands out due to its depth is likely to have depth values that abruptly change compared to its neighborhood. The depth value of the object itself is often smooth and its depth contrast is low. Visual attention is directed to objects with lower inner disparity contrast [44] but vary abruptly outside their depth value [19]. To account for this fact, a depth abruptness mask is created in which the mask contains high values when depth changes abruptly (and contains values close to zero in case depth is changing smoothly).

To create the depth abruptness mask, four points are selected around each segment (See Fig. 4 for an illustration). We observe that $P_1$ and $P_3$ have the same horizontal coordinates as the center of mass (centroid) of the segment, while $P_2$ and $P_4$ share the same vertical coordinate with the centroid. Suppose $d_1$, $d_2$, $d_3$, and $d_4$ are the highest horizontal and vertical distances from the center of mass of the segment to any point on the segment perimeter towards up, right, down, and left side directions, respectively. The four selected points are located at $P_1$:($\overline{C}_x, \overline{C}_y - d_1 - 0.01 \times H$), $P_2$:($\overline{C}_x + d_2 + 0.01 \times W, \overline{C}_y$), $P_3$:($\overline{C}_x, \overline{C}_y + d_3 + 0.01 \times H$), and $P_4$:($\overline{C}_x - d_4 - 0.01 \times W, \overline{C}_y$), where $H$ and $W$ are the height and the width of the display screen. These four points are likely to fall within neighbor segments and therefore, can be used to evaluate the depth change rate. Note that studies have shown that objects' sizes affect their visual saliency [45]. Walther et al. used a threshold of 5% for the minimum suitable area of a salient region. Our experiment settings (are explained in details in Section 3) suggest using an area threshold of 1% due to the utilized display size and viewers' distance. As a result, we add 1% of the display height to $d_1$ and $d_3$ and 1% of the display width to $d_2$ and $d_4$ to reach the four points. Next, for each segment and its associated set of selected points we find the weighted average of depth differences between the points and the centroid, and assign the resulting average value to that segment. Continuing this process for all of the segments generates a depth difference map, $Diff_{Depth}$. The weights in finding the average difference are equal to inverse of the distances between each $P_i$ point and the centroid. The depth abruptness for each segment is defined as follows:

$$Abruptness_K = \begin{cases} Max(Diff_{Depth}) & if \quad Diff^K_{Depth} > \overline{Diff_{Depth}} \\ Diff^K_{Depth} & if \quad Diff^K_{Depth} \leq \overline{Diff_{Depth}} \end{cases} \quad (13)$$

where $Diff^K_{Depth}$ represents the depth difference associated to the $K^{th}$ segment. In our formulation, we interpret a local depth difference of higher than average as an abrupt change and a depth difference of lower than average is referred to as a relatively smooth change. The final depth abruptness mask for each frame is normalized to the interval of [0-1]. Note that since synthetic depth maps can always contain artifacts, in our implementation we choose multiple (three in the present embodiment) close points instead of only one point at each direction to increase the robustness of the process against the depth map artifacts. These multiple points are selected very close (a few pixels away) to each $P_i$ point. Also, note that instead of four neighboring points, one can choose eight neighboring points (with 45º rotations), or a scanning line in each of the four directions to measure the maximum depth differences. Our simulation experiments, however, verified that the additional accuracy is negligible compared to the added computational complexity.

We perform a slight smoothing (using a simple Gaussian filter) on the depth image to prevent the imperfections that can be caused by depth map artifacts. To create the final depth saliency map, we multiply the depth abruptness mask (element-wise) by the current segmented depth map. The depth saliency map is linearly normalized to [0-1] at the end.

*7) Fovea masking*

When focusing on a specific region of an image, HVS perceives the neighborhood around that region very sharply but as the distance from the center of attention is increased the rest of the picture seems blurry to the human eye. This is due to the photoreceptor concentration density in fovea, which decreases from the center of fovea rapidly [46].

As mentioned previously in this section, we incorporate a fovea-masking-based center-surround operation in generating some of the proposed feature maps. Inspired by the photoreceptor concentration in the fovea, in our implementation, this mask is a circular disk in which the value of each element is proportional to the photoreceptor density of the corresponding location. We use the photoreceptor density values resulted from subjective biological measurements [46] to assign an approximate estimate of the photoreceptor density at each distance from the center of fovea. We use interpolation between these distances to design a continuous kernel. Radius of the mask is defined based on the size of the display and distance of the viewer from the display. Suppose α is half of the angle of the viewer's eye at the highest visual acuity. The range of 2α is between 0.5° and 2° [47]. Sharpness of vision drops off quickly beyond this range. The mask radius is approximately defined by:

$$L = Z_{observer} \tan(\alpha) \ [cm] = \frac{Z_{observer} \tan(\alpha) R_H}{H} \ [pixel] \quad (14)$$

where $H$ and $R_H$ are the vertical height and resolution of the display, and $Z_{observer}$ is the distance of the viewer to display. In our implementation (explained in Section 3), we choose an angle of α=1°, HD (High Definition) resolution video at 1080×1920, viewing distance of 183 cm, and display height of 57.25 cm. The resulting mask radius for this setup is approximately 60 pixels.

*8) Size, compactness, and sparsity of salient regions*

As already mentioned, psycho-visual experiments have revealed that object size affects the saliency [45]. In our method, segments with size of less than 1% of the resolution in each direction are not considered as salient. In addition to size of the regions, compactness of each region affects its visual saliency. Due to the nature of the eye-tracking studies (and similarly in reality), human attention is directed towards compact objects [20]. This makes more sense considering the fact in video saliency prediction there are only a few fixations per frame, which mostly are associated with compact objects. To account for the effect of object compactness, the moment of inertia for each segment is chosen as a compactness measure [48]. A map of compactness is then created in which each segment is associated with its compactness value. This map will be applied as multiplicative mask to all feature maps.

Another important factor regarding the saliency of objects is that there are only very few salient objects in each scene. Given the allocated time for human viewers to view each frame of a video and hardware capability to record their eye fixations data, there are only a couple of fixations per frame. Therefore, a proper saliency prediction algorithm should extract only a few highly probable salient regions. To account for the sparsity of the salient regions, we propose a mechanism that puts emphasis on local maxima points in the available feature maps. Assuming a feature map is scaled to [0-1], we seek a convex function that projects this feature map to another map with the desired properties. One candidate for such operation is:

$$\Phi(F) = e^F - m_F \quad (15)$$

where $F$ denotes a feature map and $m_F$ represents the average of $F$. Since $F$ values are within [0-1] then the values of Φ fall within [1-$m_F$, e-$m_F$]. For the $F$ values equal to $m_F$, Φ would approximately become 1 (normally, there are only a few salient regions in each frame. Therefore, for a sparse saliency map, $m_F$ is very small since it is equal to the average of saliency probabilities). The $F$ values above the average are subject to higher increments compared to $F$ values below the average. Once the resulting Φ-map is rescaled to [0-1], the local maxima points are relatively more amplified than the points with feature values below average. Rescaling is performed linearly as follows:

$$\Psi = \frac{\Phi - \Phi_{min}}{\Phi_{max} - \Phi_{min}} \quad (16)$$

*9) 3D Visual Discomfort*

When watching stereoscopic content, several reasons may degrade the 3D quality of experience. Assuming that the 3D display does not have any crosstalk and the data is captured properly without any unintended parallax, the main source of discomfort is caused by the vergence-accommodation conflict [49]. When viewing stereoscopic 3D content, there is a comfort zone for the content within which the objects should appear. Any region perceived outside of the comfort zone degrades the 3D QoE, as eye muscles try to focus on the display screen to perceive a sharper image while they also try to converge outside of the display screen to avoid seeing a double image. This decoupling between vergence and accommodation results in fatigue and degrades the QoE. Studies suggest using a maximum threshold for 3D content disparity. In particular as a rule-of-thumb, it is widely accepted to use maximum allowed of 1° disparity [49]. Beyond this threshold QoE drops rapidly.

Viewers tend to avoid looking at regions of a 3D scene which impose 3D visual discomfort. The visual discomfort pushes the viewers to move their gazed points away from the source of discomfort. To account for 3D discomfort, we create a discomfort penalizing filter and apply (element-wise multiplication) it to our depth feature map as it represents the 3D saliency attributes. The discomfort penalizing mask elements are assigned to 1 when a segment falls within the comfort zone and to a penalty value when the segment is outside of the comfort zone. Several quantitative measures of 3D visual discomfort are proposed so far. We don't limit our visual attention model to any particular discomfort measurement as it is not yet clear how exactly discomfort can be measured. The choice of the discomfort metric does not affect our proposed penalizing scheme. However, for the sake of illustration, we choose a simple discomfort metric, which is based on subjective visual experiments presented in [49]. By averaging the 3D QoE values of [49] for various types of content at different resolutions and disparity ranges, we derive the following rough estimate for the penalizing mask:

$$Discomfort\ Mask = \begin{cases} 1 & d \leq 60^{\text{minutes arc}} \\ 1.36 - 0.006 \times disparity & d > 60^{\text{minutes arc}} \end{cases} \quad (17)$$

In our experiment setup, $1^\circ$ disparity corresponds to 60 pixels.

*2.2 High-level saliency features*

In addition to low-level bottom-up saliency features, we add several high-level top-down features to our model. The use of high-level features helps to improve the accuracy of saliency prediction. The following high level features are considered in our method: face, person, vehicle, animal, text, and horizon. For each feature, a saliency map (using a bounding box around the detected salient region) is created and used in the training of the proposed visual attention model.

When a human appears in a video shot, the observer's attention is naturally drawn to the person. In order to detect faces and appearance of people in a scene, we use the Viola-Jones algorithm [50] and Felzenszwalb's method [51] (trained on the PASCAL VOC 2008 dataset [52]), respectively. Felzenszwalb's method [51] is also used to detect the presence of bicycles, motorbikes, airplanes, boats, buses, cars, and/or trains. The same method is incorporated to detect animals including birds, cats, cows, dogs, horses, and/or sheep.

Image areas containing text also attract the human attention. To account for the appearance of text we use the Tesseract OCR (Optical Character Recognition) engine [53]. In addition, Gist descriptor is used to detect a horizon in the scene, which itself is shown to be potentially a salient feature [54].

Fig. 5, demonstrates generated feature maps for a sample video frame. Note that the feature maps provided in Fig. 5 are just samples generated from a sample frame. The overall performance of the features are evaluated in Section 4.

*2.3 Feature map fusion based on random forests*

Since it is not clear how map fusion happens in the brain, computational models of visual attention use different approaches. On one hand, some approaches integrate the features internally and do not produce several separate feature maps [8-10,15,16]. One the other hand, other methods incorporate linear [5-6,12-13,14], SVM (support vector machine) [7], or deep-networks-based map fusion schemes [55]. In the case of 3D visual attention modeling, map fusion is mostly performed by linear combination [18-20].

In our approach, we utilize random forest regression for fusing the different feature maps generated by our method. Random forests regression is an ensemble learning method that constructs a multitude of decision trees at training time and outputs the mean prediction (regression) of the individual trees. Random decision forests correct for decision trees over-fitting problem. More details regarding random forests can be found in [86].

The motivation behind this tool selection is explained below. Random Forest (RF) methods construct a collection of decision trees using random selection of input features samples [56]. Generally, each decision tree might not perform well over unseen data. However, the ensemble of the trees usually generalizes well for test data. This follows the structure of HVS (and also the proposed feature extraction process) in incorporating several visual features. Each feature on its own may not predict the saliency well, but integrating various features provides a much more accurate prediction. One of advantages of RF regression techniques is that they do not require extensive parameter tuning since they intuitively divide data over the trees based on how well they classify the samples. Moreover, the importance of each individual feature can be evaluated using the out-of-bag-errors once a model is trained. This makes it possible for our proposed method to find an assessment of how important each feature is in the saliency prediction, so that an appropriate decision can be made on the trade-off between number of features (complexity) and prediction accuracy. Note that other learning frameworks can be substituted with random forest. However, due to reasons mentioned above, we utilize random forests approach in our method. As we will see in section 4, using random forests for feature fusion shows a promising performance, superior to some existing alternative methods.

It is also worth noting that random forest involves with sampling the training data (bootstrapping) with replacement. In this sampling, a percentage of the data (usually 30%) is not used for training and can be used to testing. These are called out-of-bag samples. Error estimated on these out of bag samples is the out-of-bag-error, which has proven to be unbiased in many tests [72].

For a set of training videos (details in Section 3), we extract different feature maps and use them to train a RF regression model. Bagging is applied to tree learners to construct the decision trees. The resulting RF model is later used to generate a saliency map for unseen test features. In addition, the importance of each feature helps to decide what number of features to use.

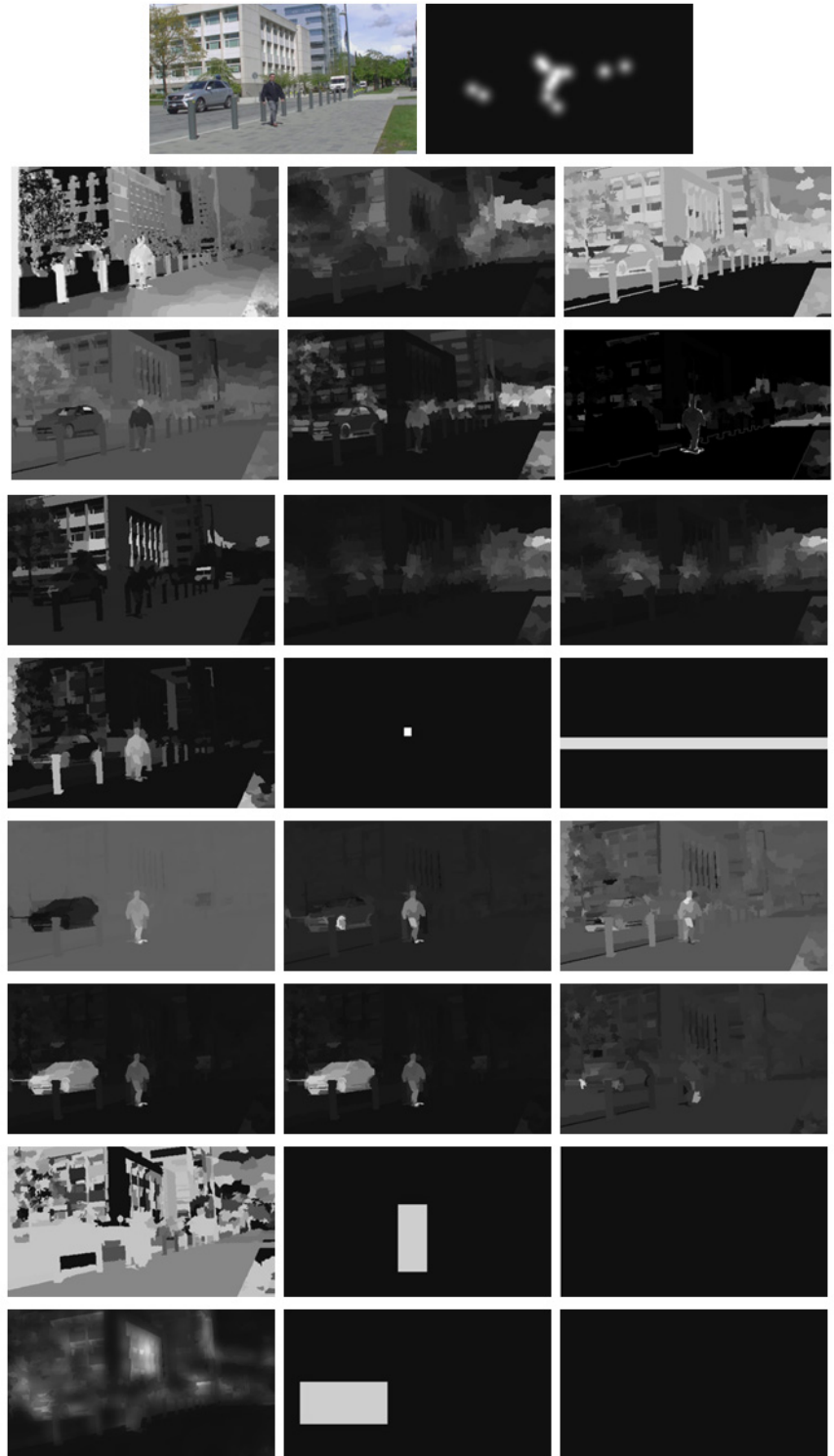

**Fig. 5** Generated feature maps using the proposed method: Images from left to right in a raster-scan order are: Right view picture, human fixation map, disparity map, brightness feature, color histogram feature, color contrast feature of a$^*$, color contrast feature of b$^*$, color warmth, color saturation, HVS color sensitivity feature, empirical color saliency feature, depth feature, face, horizon, motion *dx*, motion *dy*, motion *dz*, velocity magnitude, velocity magnitude with emphasize on *Z* direction, acceleration, element of surprise, person, text, texture, vehicle, animals

**3 Experiment Settings**

This section provides a brief overview about our stereoscopic video database and subjective experiments. Complete details regarding the video database as well as the eye-tracking experiments can be found in [57] where we introduce our benchmark saliency prediction database.

*3.1 Stereoscopic video database capturing*

To this date (to the best of our knowledge), the only publicly available stereoscopic video databases for saliency prediction experiments are the EyeC3D [58] and the IRCCyN [59] datasets, which contain 8 and 47 stereoscopic videos, respectively. To create a large-scale stereoscopic video database with ground-truth eye-tracking data, we captured 61 indoor/outdoor sequences with a wide range of depth, motion, brightness, and texture density. There is roughly an equal number of scenes with and without humans. Similarly, we tried to include an equal number of scenes with and without moving vehicles. In general, it is ensured that the database includes scenes with a variety of different combinations of objects of interest in front or at the back of the screen, high and low brightness, fast, medium and slow motion, dense and sparse texture, with and without vehicles and/or humans. The length of each video sequence is approximately 10 seconds.

*3.2 Post-processing the captured videos*

Disparity correction was performed to bring the objects of the interest on the display screen (within the 3D viewing comfort zone). This is common practice, as studies have shown that this is preferable compared to letting objects appear in front of the screen [60] (See [41] for more details).

*3.3 Subjective experiments*

Eye fixation points of the participants are tracked using a SMI iView X RED device [61]. The eye-tracker is placed between the subjects and a 3D TV in a way that the requirements of the SMI system are met. The sampling frequency of the SMI is 250 Hz and the resolution accuracy is $0.04 \pm 0.03^{o}$. To display the test material, a 46" Hyundai S465D 3D TV with passive glasses is used. Resolution of the display is the same as the video sequences (HD, 1080×1920), avoiding rescaling the video content on the screen.

A total of 24 subjects participated in our experiments, with ages ranging from 20 to 30. Having 61 videos each around 10 seconds long, accumulated to about 10 minutes of actual test time per each subject. All participants were screened for visual acuity (Snellen charts), color blindness (Ishihara chart), and stereovision acuity (Randot test) to ensure they were eligible to do the test. Moreover, they were naïve regarding the test purpose. Each test session was a free-viewing task, i.e., subjects were just asked to view the video sequence without any specific task to do. To ensure accuracy of eye tracking, a calibration step was performed for each subject several times, at the beginning and throughout the test so that the eye-tracker did not lose the track of the eye movements. More details regarding the subjective experiments can be found in [57].

Once the gazed point data for all subjects are collected, eye-fixation maps are created from the gazed points by using a Gaussian filter. As suggested by the state-of-the-art [7,17,18,29,57], we use a Gaussian filter with one degree (of visual angle) standard deviation.

Our database of stereo videos as well as the eye-tracking data are available at http://dml.ece.ubc.ca [57].

**4 Results and Discussions**

This section elaborates on the results of our experiments and compares the performance of the proposed saliency prediction method with that of the state-of-the-art.

*4.1 Metrics of performance*

A primary metric of performance in our analysis is the Receiver Operating Characteristics (ROC) and the Area Under the ROC Curve (AUC) [62]. The classic definition of ROC imposes a bias towards the center, which can significantly affect the accuracy of the AUC metric [16,62]. Shuffled AUC (sAUC) is a modified version of AUC, which tackles this issue [16,62] and provides more robustness.

Other than the AUC and sAUC, in our performance evaluations we use the Earth Mover's Distance (EMD) metric [63] to account for the spatial distance in the saliency maps (not only the ordering), Kullback-Leibler Divergence (KLD) [64], Normalized Scanpath Saliency (NSS) [62], Pearson Correlation Coefficient (PCC), as well as a similarity measure, SIM, proposed by Judd et al. [65]. Note that except KLD and EMD which are distance metrics, the higher the metric values are, the better the performance is.

**Table 1.** Relative Feature Importance (RFI)

| Feature | RFI | Feature | RFI |
|---|---|---|---|
| Motion2 (Dy) | 1 | Face | 0.67 |
| Person | 0.92 | Motion4 (V) | 0.61 |
| Brightness Var. Contrast | 0.89 | Color3 (Saturation) | 0.57 |
| Color4 (HVS sensitivity) | 0.88 | Motion7 (Surprise element) | 0.55 |
| Depth | 0.85 | Color1 (histogram) | 0.52 |
| Motion1 (Dx) | 0.84 | Color7 (Contrast-a′) | 0.52 |
| Brightness Contrast | 0.80 | Color6 (Contrast-b′) | 0.52 |
| Color2 (Warmth) | 0.78 | Motion5 (Z-emphasis) | 0.47 |
| Motion6 (A) | 0.70 | Color5 (Empirical) | 0.39 |
| Motion3 (Dz) | 0.70 | Text | 0.25 |
| Texture | 0.69 | Horizon | 0.18 |
| Vehicle | 0.68 | Animals | 0 |

*4.2 Contribution of each proposed feature map*

A total of 24 sequences were selected for training the random forest model and the rest (37 videos) were used for performance validation. We categorized the videos in a way that both the training and validation sets contain videos with a wide variety of possible scenarios of depth range, brightness, motion, etc. The training set is used to train a random forest model with all 24 features. We chose 40 trees for training the random forest and around 7% of the training data (540 frames) for fast implementation. The impact of learning parameters is investigated in the next subsections. The resulting model achieves AUC=0.7243, sAUC=0.7795, EMD=0.4528, SIM=0.2966, PCC=0.2620, KLD=0.1289, and NSS=1.4167.

The use of random forest ensemble learning makes it possible to extract the relative importance of each feature compared to other ones by comparing their corresponding "out of bag" error values. Table 1 shows the relative feature importance values. We can observe that among the high-level features, the presence of humans is of the highest importance. Among the low-level features, motion, brightness, depth, and color are the most important ones in saliency prediction. Note that since there are no animals in the scenes, the importance value for this feature is equal to zero. Due to the flexibility of the incorporated learning method, any of the existing features can be removed from the model, or new features can be added.

Fig. 6.(a)-(c) show the performance metrics as functions of the number of features, when the first $i$ ($i=1,2,...,24$) important features are used. The results in this figure verify that higher performance is achieved by using a larger number of features. However, higher number of features increases the complexity for both training and validation. An analysis of complexity is provided in subsection 5. It is worth mentioning that according to Fig. 6.(a)-(c), using only the first 12 important features (half of the features) results in 95.5% percent (in terms of sAUC) of the highest algorithm performance (when the entire feature set is used).

We also evaluated the saliency detection performance of only the low-level features, i.e., the ones extracted from brightness, color, texture, motion, and depth. This helps us understand the influence of low-level and high-level features in visual attention. To this end, a model was trained using the low-level features and was tested using the validation set. The resulting metric values are 0.7019, 0.7565, 0.5107, 0.2663, 0.2529, 0.1505, and 1.2948 for AUC, sAUC, EMD, SIM, PCC, KLD, and NSS, respectively. The resulting values are close to the ones corresponding to the proposed model with the entire feature set. This shows that while our model performs fairly well using only the low-level features, it achieves its highest performance when all features are used. It is worth noting that using a larger number of features results in higher computational complexity. The proposed framework provides flexibility for users to select a suitable point of trade-off between the accuracy and complexity. An analysis of complexity is provided in subsection 5.

*4.3 Tuning the training parameters*

In this subsection we study the impact of each parameter in the model performance.

1) Size of the training data: Using a different number of frames from each training video may change the learning performance. The training video dataset consists of 24 stereoscopic videos. In the present implementation, we pick the first 20-to-25 frames of each video (540 frames in total). To investigate the impact of the size of the training data, we select different number of frames from each sequence to train new models. Performance evaluations show that, in general, even using a small portion of the training dataset results in an acceptable accuracy and that the saliency prediction accuracy is not highly sensitive to the size of the training dataset. Fig. 6.(d)-(f) show the performance of our VAM for different sizes of the training data.

2) Random forest parameters: In the ensemble random forest learning method, we used boot strapping with sample ratio of 1/3. The minimum number of observations per tree leaf is set to 10. In the current implementation, the number of trees is set to 40 for a fast performance. To study the impact of the number of trees on the saliency prediction, we train additional models with different number of trees, and evaluate their performances over the validation video set. We observe that choosing different number of trees, between 1 and 100, results in very smooth variations in the performance metrics, with slight improvement as the number of trees increases. However, choosing a very large number of trees possibly results in over-fitting and may degrade the overall performance (See Fig.6.(g)-(i)).

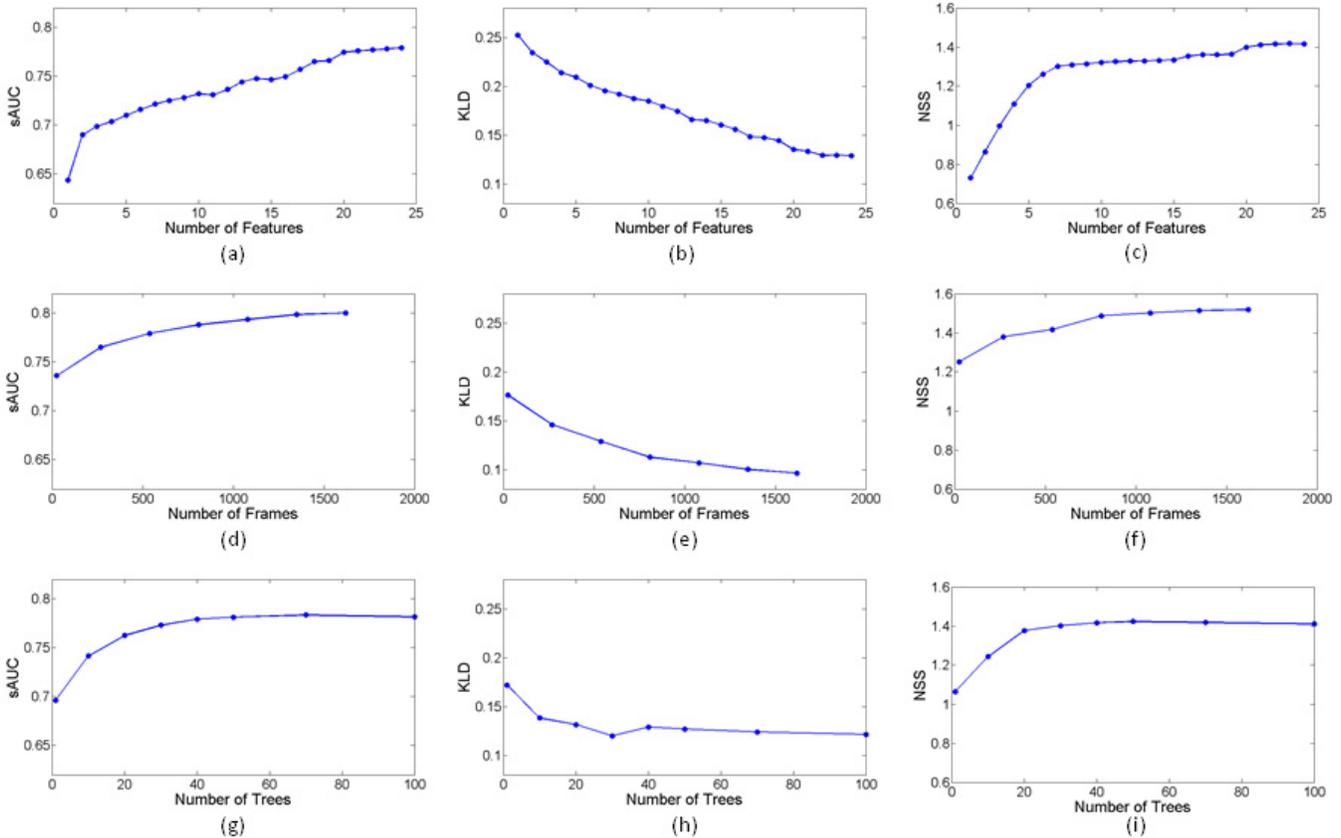

**Fig. 6.** Performance metrics when different number of features are used: AUC (a), sAUC (b), PCC (c), NSS (d), SIM (e), KLD (f), EMD (g).

*4.4 Comparison with different map fusion approaches*

It is common practice to fuse various conspicuity maps into a final saliency map. The proposed random forest approach combines the individual maps efficiently and according to their relative importance. To demonstrate the strength of random forests in map fusion, we provide a comparison between the performance of our model and different fusion schemes adopted in the state-of-the-art methods. In particular, we compare our method against: 1) Averaging (finding the average of different conspicuity/feature maps), 2) Multiplication, 3) Maximum, 4) Sum plus Product (SpP), 5) Global Non-Linear Normalization followed by Summation (GNLNS) [11], 6) Least Mean Squares Weighted Average (LMSWA), 7) Standard Deviation Weight (SDW), and 8) Support Vector Regression (SVR).

**Table 2.** Evaluation of Different Feature Fusion Methods

| Fusion Method | AUC | sAUC | EMD | SIM | PCC | KLD | NSS |
|---|---|---|---|---|---|---|---|
| Average | 0.677 | 0.714 | 0.780 | 0.222 | 0.195 | 0.218 | 1.116 |
| Multiplication | 0.555 | 0.569 | 1.109 | 0.177 | 0.168 | 1.020 | 0.877 |
| Maximum | 0.594 | 0.586 | 1.001 | 0.190 | 0.196 | 0.899 | 0.997 |
| SpP | 0.667 | 0.698 | 0.898 | 0.236 | 0.210 | 0.205 | 1.118 |
| GNLNS [11] | 0.703 | 0.757 | 0.809 | 0.221 | 0.259 | 0.688 | 1.299 |
| LMSWA | 0.692 | 0.744 | 0.559 | 0.268 | 0.247 | 0.188 | 1.295 |
| SDW | 0.657 | 0.691 | 0.786 | 0.224 | 0.199 | 0.193 | 0.995 |
| SVR | 0.709 | 0.758 | **0.439** | 0.270 | 0.219 | 0.166 | 1.330 |
| **Random Forest** | **0.724** | **0.780** | 0.453 | **0.297** | **0.262** | **0.129** | **1.417** |

Table 2 shows the results of these fusion methods and our random forest approach, clearly indicating that our fusion outperforms the other types of map fusion.

*4.5 Computational complexity*

As mentioned in the previous sub-sections, the proposed learning-based framework provides the flexibility to add/remove additional saliency features. While some of the features are easier to extract, some might introduce more overall computational complexity. The present embodiment uses 24 low and high level features, however, depending on the type of scenes and desired accuracy and computational cost, users may choose different number of features to train the model. Fig. 7 shows the simulation runtime for our model when different number of features is used. To generate this figure, we used the model parameters mentioned in sub-section 4.2 (540 training frames from 24 video sequences, and 40 trees). Order of the features is according to Table 1, starting from most important feature to the least important one. Note that since training is performed only once, a generated model can be saved and applied on test datasets. As a result, times reported in this figure, include average time, for a 10 second (30 fps) video in test database, that takes to generate features and fuse them to a saliency map (for all frames).

Note that the mathematical definition of the complexity of an algorithm involves calculating the number of operations (e.g. using big O notation). Due to the complex structure of most of the visual attention models, and lack of implementation details in some of the associated papers, it is not practical to calculate their mathematical complexity. The feasible solution to compare the complexity of different algorithms is therefore to compare their simulation times. The runtime measurements reported in Fig. 7 provide a sense of computational complexity for each algorithm (but not the true algorithm complexity). That being said, for our runtime measurements we used a workstation with i7 CPU and 18 GBs of memory to perform complexity measurements. Note that it was ensured that no other background process was running during the measurement process. It is also worth noting that some of the algorithms have a parallel nature and thus can be boosted by efficient GPU implementations.

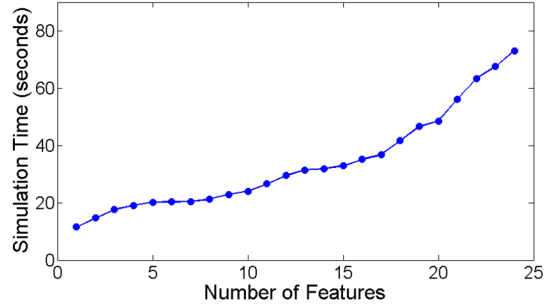

**Fig. 7.** Complexity (execution time) when different number of features is used. Note that the present implementation is serial however parallel implementation is possible both in feature extraction and RF fusion.

*4.6 Comparison with the state-of-the-art VAMs*

In this subsection, we provide a comparison between the performance of the proposed saliency prediction method and the state-of-the-art visual attention models, which were originally proposed for saliency detection on 3D images and videos. In addition, four baseline models (representing trivial cases) are generated and added to the comparisons, a common practice in VAM performance evaluations.

**Baselines**: The baselines are as follow:
1) Chance: a map of random values in [0-1].
2) Center: Center map is a Gaussian circular disk placed at the center of the image.
3) One human: Human observers do not have the exact same opinion when watching a scene. Therefore, fixations are spread around the salient regions. Similar to the MIT saliency benchmark [65], we create a baseline of the fixation maps generated by only one human and compare this map with the average fixation map of the other observers. To this end, the fixation map of each participant is used as a saliency map and its performance in predicting the fixation maps of other participants is evaluated. This process is repeated for all the subjects and the average performances are used for the baseline.
4) Infinite humans: we use the statistics of the 24 participants to find an estimate of a saliency map, which would have been produced by an infinite number of humans. Details regarding the baseline models can be found in [57].

**Blurring and center-bias**: It is a common practice to add a center-bias and blur a saliency map before evaluating its performance using objective metrics [17,65]. We linearly add a center-bias and perform the blurring (using a Gaussian kernel) for our saliency maps as well as the other VAMs. Standard deviation of the center map and the Gaussian kernel and the weight for the center prior are assigned for each VAM separately, in a way that provides the maximum AUC with the training video set [57]. In addition, to ensure a fair comparison between different VAMs, we apply histogram matching to match the histogram of the saliency maps with their corresponding fixation maps [65].

**Table 3.** Performance Evaluation of Different VAMs Using Our Eye-Tracking Dataset of Stereoscopic Videos

| Model | AUC | sAUC | EMD | SIM | PCC | KLD | NSS | Simulation Time (sec) | Average Rank* | Type |
|---|---|---|---|---|---|---|---|---|---|---|
| Infinite humans | 0.9921 | 0.9908 | 0.03 | 0.9511 | 0.9968 | 0 | 4.2524 | | 1 | |
| LBVS-3D (proposed) | 0.7243 | 0.7795 | 0.4528 | 0.2966 | 0.2620 | 0.1289 | 1.4167 | 73.39 | 2.33 | 3D video |
| LBVS-3D (static**) | 0.6833 | 0.7091 | 0.5310 | 0.2544 | 0.2376 | 0.1963 | 1.1782 | 30.55 | 3.33 | 3D image |
| One human | 0.7033 | 0.7379 | 0.8884 | 0.4651 | 0.4995 | 0.2232 | 2.1140 | | 5.33 | |
| Fang [17] | 0.6655 | 0.6915 | 0.6676 | 0.2229 | 0.1987 | 0.2165 | 1.0380 | 3.25 | 5.33 | 3D image |
| Coria [67] | 0.6584 | 0.6843 | 0.6568 | 0.2346 | 0.1417 | 0.2238 | 1.1361 | 3.03 | 8 | 3D video |
| Chamaret [27] | 0.6669 | 0.6787 | 0.7568 | 0.2089 | 0.1568 | 0.2253 | 0.9056 | 64.62 | 8.33 | 3D video |
| Park [68] | 0.6391 | 0.6346 | 0.8081 | 0.1841 | 0.1022 | 0.2198 | 0.7783 | 1.68 | 9 | 3D image |
| Ouerhani [28] | 0.6224 | 0.6456 | 0.8768 | 0.1934 | 0.0967 | 0.2179 | 0.5459 | 7.21 | 9.33 | 3D image |
| Fan [69] | 0.6349 | 0.6330 | 0.9014 | 0.1879 | 0.0856 | 0.2116 | 0.4185 | 128.98 | 9.33 | 3D image |
| Niu [19] | 0.6078 | 0.6124 | 0.9339 | 0.1726 | 0.1208 | 0.2227 | 0.3334 | 165.82 | 9.67 | 3D image |
| Ju [70] | 0.5811 | 0.5948 | 1.0330 | 0.1623 | 0.0827 | 0.2109 | 0.2778 | 2.05 | 10.67 | 3D image |
| Jiang [71] | 0.6158 | 0.6089 | 0.9949 | 0.1934 | 0.1211 | 0.2326 | 0.3656 | 1.25 | 11.33 | 3D image |
| Center | 0.5709 | 0.5999 | 0.6536 | 0.2128 | 0.1104 | 0.2445 | 0.6524 | 0.06 | 13 | |
| Zhang [26] | 0.5699 | 0.5754 | 1.0970 | 0.1528 | 0.0563 | 0.2293 | 0.2111 | 0.73 | 14.33 | 3D image |
| Chance | 0.5 | 0.5 | 1.1140 | 0.1421 | 0 | 0.2393 | 0.0789 | 0.072 | 15.67 | |

\* Ranking is done using only sAUC, KLD, and NSS.  \*\* Motion features are excluded.

The performance evaluations are provided in Table 3. In order to sort different models according to their performance, we assign a separate rank for each metric and use the average ranks. It was shown in [66] that most performance metrics are similar and that using sAUC, KLD, and one other metric among NSS, PLCC, SROCC, and SIM provides a fair comparison. To avoid the introduction of bias in the ranking, we use only sAUC, KLD, and NSS for ranking the performance of various VAMs.

Fig. 8 demonstrates a visual comparison between various VAMs. For illustration purposes, we show a left view frame from one of the videos, associated depth map, ground truth from eye-tracking, as well as saliency maps produced out of different algorithms. Note that this figure only shows one sample frame for demonstration.

To compare the performance of different algorithms, it is also essential to evaluate how these algorithms are statistically different. The metric evaluations values reported in Table 3 are average values over the entire database, however, we need to know the performance distribution across the individual videos. To this end, we compute the 95% confidence interval for each algorithm when performance is evaluated by comparing algorithm saliency maps and human fixations maps for individual videos. The confidence intervals (when sAUC, KLD, and NSS metrics are used for performance evaluations) are illustrated in Fig. 9. In addition, in case of interval overlaps, we measure the statistical difference between distributions of metric values over videos, by measuring P-values form Student T-test. Fig. 10 shows the P-values among different algorithms. The empty elements of Fig. 10 represent a zero P-value (i.e. two statistically different distributions).

Comparing the result of Table 3 with those of the MIT saliency benchmark [65] and also [57] confirms that saliency prediction in 3D is a much more difficult task than 2D, as the models provide much higher performances for the 2D case [65],[57]. The same conclusion can be made by comparing the performance of one human in predicting the saliency of an infinite number of humans. The fact that one human is not a good representative for infinite number of humans (according to Table 3) is due to the complex structure of 3D perception and the additional saliency attributes introduced in 3D.

It is worth noting that high level features (and all other features too) are combined based on their importance. The combining process (feature fusion) takes place inside the random forest module. If there are multiple high level features in a scene, some are likely to be looked at, and some are not. Based on the eye-tracking training data available to us from subjective experiments, we can learn how viewers preferred one high level object to the others. In fact, this is how feature importance values are calculated. For instance, according to Table 1, appearance of people in a scene is much more salient than appearance of vehicles. As a result the learnt model will assign higher saliency values to locations in a scene that people are present. In short, handling multiple high level features in a scene is based on the eye-tracking data that is learnt. This brings a limitation to this approach that there are only many scenarios that can be available as training data, so not every situation can be truly learnt. On the positive side however, due to flexibility of our approach, one can train a model with desired types of scenes, so that the method achieves higher accuracy on similar target videos. Moreover, flexibility of our approach allows one to add or remove any type of high or low level feature. The ones introduced in our implementation are just sample suggested features that showed potential in saliency prediction.

Advantages of our saliency detection approach includes: having an acceptable accuracy, using both high and low level features, flexibility of the model to add or remove features, and often being reliable for different scene types (because it's trained with various scene types). A disadvantages of our approach is that generation of all features can be time consuming (see Table 3). Also, it might not be possible to extract high fidelity high-level features such as face or vehicles due to challenging scene contents.

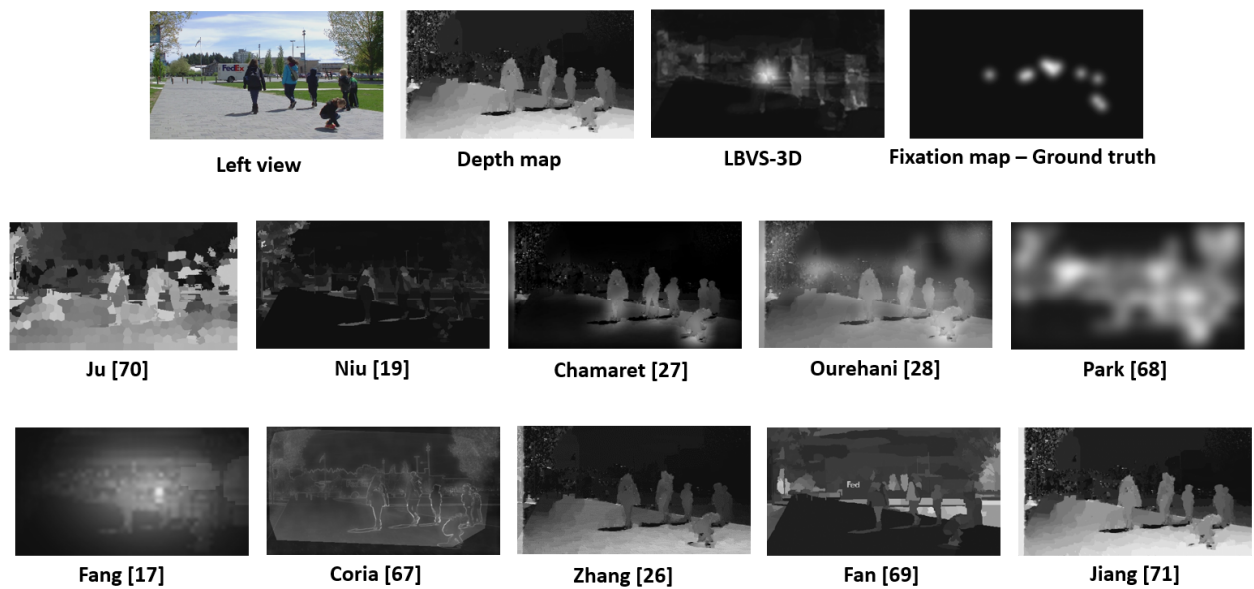

**Fig. 8.** A demonstration of saliency prediction of various stereoscopic VAMs

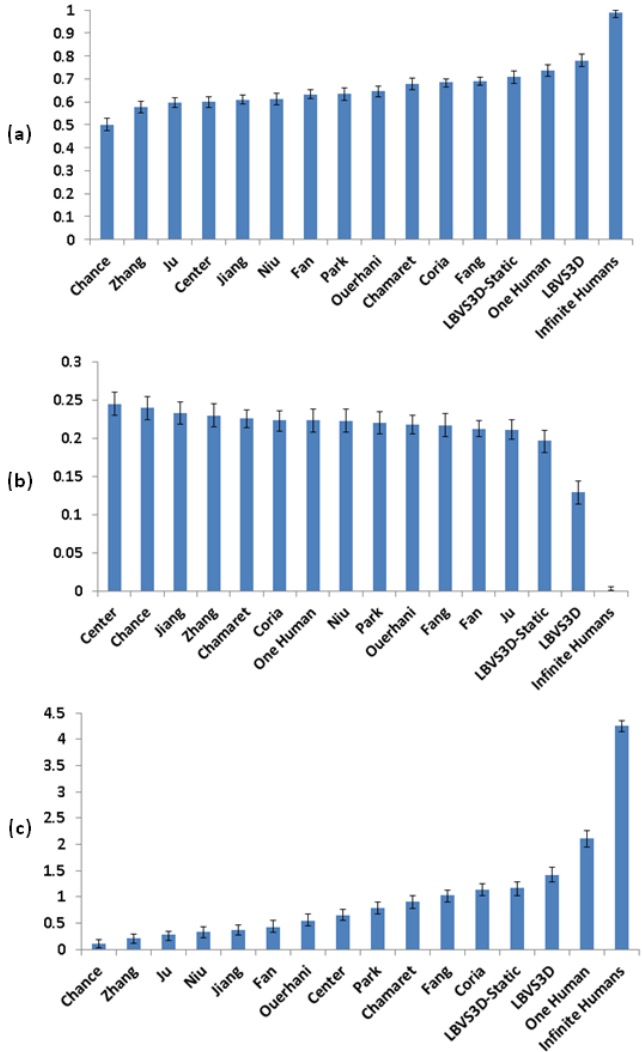
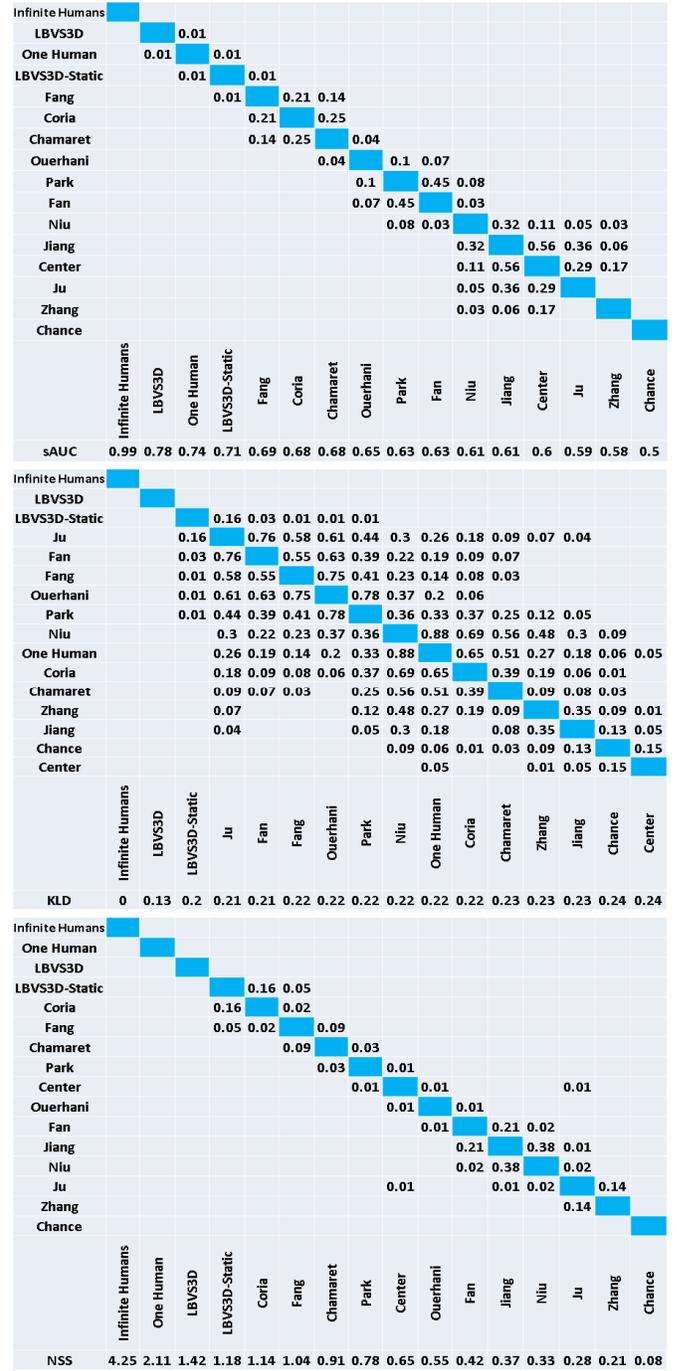

**Fig. 9.** Performance of different VAMs with their 95% confidence intervals: (a) sAUC, (b) KLD, and (c) NSS

**Fig. 10.** P-Values from student T test between each pair of VAMs: (top) sAUC, (center) KLD, and (bottom) NSS

*4.7 Comparison with the state-of-the-art VAMs – using another public dataset*

To ensure that the performance evaluations provided in Section 4 are not biased towards any particular type of video, we provide a second set of performance evaluations using another publicly available stereo video database. To this end, we utilize the database introduced in reference [59], which consists of eye-tracking data of 41 scenes, at 1920 × 1080 resolution, from 40 subjects. Table 4 summarizes the performance of various VAMs when incorporated for saliency prediction on this database. The same metrics and ranking method as previous sub-sections is used. Despite some minor variations, results are in general in agreement with Table 3.

Table 4. Performance Evaluation of Different VAMs Using Eye-Tracking Dataset of [59]

| Model | AUC | sAUC | EMD | SIM | PCC | KLD | NSS | Average Rank* | Type |
|---|---|---|---|---|---|---|---|---|---|
| LBVS-3D (proposed) | 0.74 | 0.78 | 0.40 | 0.32 | 0.26 | 0.10 | 1.55 | 1 | 3D video |
| Coria [67] | 0.68 | 0.70 | 0.58 | 0.27 | 0.23 | 0.18 | 1.06 | 2.3333 | 3D video |
| Fang [17] | 0.679 | 0.688 | 0.55 | 0.28 | 0.22 | 0.19 | 1.18 | 2.6667 | 3D image |
| Chamaret [27] | 0.671 | 0.684 | 0.61 | 0.26 | 0.205 | 0.200 | 1.02 | 4.3333 | 3D video |
| Park [68] | 0.665 | 0.675 | 0.68 | 0.246 | 0.18 | 0.208 | 1.05 | 5 | 3D image |
| Ouerhani [28] | 0.6648 | 0.680 | 0.75 | 0.241 | 0.14 | 0.222 | 1.00 | 6.3333 | 3D image |
| Niu [19] | 0.6640 | 0.672 | 0.72 | 0.243 | 0.200 | 0.211 | 0.98 | 6.6667 | 3D image |
| Fan [69] | 0.6146 | 0.64 | 0.86 | 0.22 | 0.13 | 0.216 | 0.67 | 7.6667 | 3D image |
| Ju [70] | 0.60 | 0.63 | 0.88 | 0.18 | 0.11 | 0.229 | 0.56 | 9 | 3D image |
| Jiang [71] | 0.61 | 0.62 | 0.93 | 0.173 | 0.10 | 0.23 | 0.46 | 10 | 3D image |
| Zhang [26] | 0.58 | 0.60 | 0.97 | 0.172 | 0.086 | 0.24 | 0.31 | 11 | 3D image |

* Ranking is done using only sAUC, KLD, and NSS.

## 5 Conclusion

This paper introduces a new computational visual attention model (VAM) for stereoscopic 3D video. Both low and high level features are incorporated in the design of our model. Several intuitive biological observations are quantified and adopted in our method. A random forest learning algorithm is utilized to train a saliency prediction model and efficiently fuse various feature maps generated by the proposed approach. Our method is flexible in that it allows new features to be added without changing the structure of the model. To verify the performance of the proposed VAM, we capture a dataset of stereoscopic videos and collect their eye-tracking results. Performance evaluations demonstrated the high performance of our visual attention model.

Future works include the investigation of potential usage of saliency prediction to improve the performance of 3D video quality metrics [73-77], and similarly for HDR or 3D-HDR metrics [78-80].

## 6 References

[1] J. Wolfe, "Visual attention," *Seeing*, Academic, 2000, pp. 335-386.
[2] U. Neisser, *Cognitive Psychology*. Appleton-Century-Crofts, 1967.
[3] A Treisman and G. Gelade, "A feature integration theory of attention," *Cognitive Psychology*, vol. 12, pp. 97-136, 1980
[4] K. Duncan and S. Sarkar, "Saliency in images and videos: a brief survey," *IET Computer Vision*, vol. 6, pp. 514-523, 2012.
[5] L. Itti, C. Koch, "A saliency-based search mechanism for overt and covert shifts of visual attention," *Vision Research*, vol. 40, May, 2000.
[6] J. Zhang and S. Sclaroff, "Saliency detection: a boolean map approach," ICCV 2013.
[7] T. Judd, K. Ehinger, F. Durand, and A. Torralba, "Learning to predict where humans look," ICCV 2009.
[8] E. Erdem and A. Erdem, "Visual saliency estimation by nonlinearly integrating features using region covariances," *Journal of Vision*, 2013.
[9] H. Tavakoli, E. Rahtu, J. Heikkila, "Fast and efficient saliency detection using sparse sampling and kernel density estimation," SCIA 2011.
[10] J. Harel, C. Koch, and P. Perona, "Graph-Based Visual Saliency," NIPS 2006.
[11] L. Itti, C. Koch, and E. Niebur, "A model of saliency-based visual attention for rapid scene analysis," *IEEE TPAMI*, vol. 20, Nov 1998.
[12] Y. Zhai and M. Shah, "Visual attention detection in video sequences using spatiotemporal cues," in *ACM Multimedia*, 2006, pp. 815–824.
[13] N. Riche et al., "A multi-scale rarity-based saliency detection with its comparative statistical analysis," *Sig. Proc. Img. Comm.*, 2013.
[14] S. Goferman, L. Zelnik-Manor, and A. Tal, "Context-aware saliency detection," CVPR 2010.
[15] H.J. Seo and P. Milanfar. "Static and space-time visual saliency detection by self-resemblance," *Journal of Vision*, 2012.
[16] L. Zhang et al., "SUN: A Bayesian framework for saliency using natural statistics," *Journal of Vision*, 2008.
[17] Y. Fang et al., "Saliency detection for stereoscopic images," *IEEE TIP*, vol. 23, no. 6, pp. 2625-2636, Feb. 2014.
[18] J. Wang et al., "Computational model of stereoscopic 3D visual saliency," *IEEE TIP*, vol. 22, June 2013.
[19] Y. Niu et al., "Leveraging stereopsis for saliency analysis," CVPR 2012.
[20] H. Kim, S. Lee, and A.C. Bovik, "Saliency prediction on stereoscopic videos," *IEEE TIP*, vol. 23, no. 4, April 2014.
[21] P. Seuntiens, "Visual Experience of 3D TV," Doctoral thesis, Eindhoven University of Technology, 2006.
[22] W. Chen, J. Fournier, M. Barkowsky, and P. Le Callet, "Quality of Experience Model for 3DTV," SPIE, San Francisco, USA, 2012.
[23] A. Boev et al., "Classification and simulation of stereoscopic artefacts in mobile 3DTV content," Electronic Imaging Symposium, Jan. 2009.
[24] L. Coria, D. Xu and P. Nasiopoulos, "Quality of Experience of Stereoscopic Content on Displays of Different Sizes: a Comprehensive Subjective Evaluation," IEEE ICCE, January 9-12, 2011, pp. 778-779.
[25] A. Maki, P. Nordlund, and J. Eklundh, "A computational model of depth based attention," IEEE 13[th] Int. Conf. Pattern Recognition., Aug. 1996.
[26] Y. Zhang et al., "Stereoscopic visual attention model for 3D video," *Advances in Multimedia Modeling*, Springer-Verlag, 2010, pp. 314–324.
[27] C. Chamaret, S. Godefroy, P. Lopez, and O. L. Meur, "Adaptive 3D rendering based on region-of-interest," in Proc. *SPIE*, Feb. 2010.
[28] N. Ouerhani and H. Hugli, "Computing visual attention from scene depth," IEEE 15th Int. Conf. Pattern Recognition, Sep. 2000.
[29] C. Lang, et al., "Depth Matters: Influence of Depth Cues on Visual Saliency," ECCV 2012.
[30] M. Tanimoto, T. Fujii, and K. Suzuki, "Video depth estimation reference software (DERS) with image segmentation and block matching," ISO/IEC JTC1/SC29/WG11 MPEG/M16092, Switzerland, 2009.
[31] D. C. Bourassa, I. C. McManus, and M. P. Bryden, "Handedness and eye-dominance: a meta-analysis of their relationship," *Laterality*, vol. 1, no. 1, 1996, pp. 5-34.
[32] D. Comanicu and P. Meer, "Mean shift: A robust approach toward feature space analysis," *IEEE PAMI,* vol. 24, pp. 603-619, May 2002.
[33] F.M. Adams and C.E. Osgood, "A cross-cultural study of the affective meanings of color," *Journal of Cross-Cultural Psychology*, vol. 4, 1973.
[34] E.F. Schubert, "Light emitting diodes," 2[nd] Ed. Cambridge University Press, 2006.
[35] M. Baik et al., "Investigation of eye-catching colors using eye tracking," Proc. of SPIE-IS&T Electronic Imaging, SPIE vol. 8651, 2013.

[36] M. Tian, S. Wan, and L. Yue, "A Color saliency model for salient objects detection in natural scenes," Advances in Multimedia Modeling, Lecture Notes in Computer Science vol. 5916, 2010, pp 240-250.
[37] E. Lübbe, "Colours in the mind - colour systems in reality," 2010.
[38] T. Erdogan, How to calculate luminosity, dominant wavelength, and excitation purity," Semrock White Paper Series.
[39] E.D. Gelasca, D. Tomasic, and T. Ebrahimi, "Which colors best catch your eyes: a subjective study of color saliency," ISCAS 2005.
[40] M. Drulea and S. Nedevschi, "Motion estimation using the correlation transform," *IEEE TIP*, vol.22, no.8, pp.3260-3270, Aug. 2013.
[41] A. Banitalebi-Dehkordi, M.T. Pourazad, and Panos Nasiopoulos, "The effect of frame rate on 3D video quality and bitrate," *Springer Journal of 3D Research*, vol. 6:1, pp. 5-34, 2015, DOI 10.1007/s13319-014-0034-3.
[42] A. Banitalebi-Dehkordi, M.T. Pourazad, and Panos Nasiopoulos, "Effect of high frame rates on 3D video quality of experience," International Conference on Consumer Electronics, ICCE 2014.
[43] J.Y. Lin, S. Franconeri, and J.T. Enns, "Objects on a collision path with the observer demand attention," *Psychology Science*, vol. 19, 2008.
[44] Y. Liu, L.K. Cormack, and A.C. Bovik, "Natural scene statistics at stereo fixations," 2010 Symposium on Eye-Tracking Research & Applications, ETRA 2010.
[45] K.I. Beverley and D. Regan, "Visual perception of changing size: The effect of object size," *J. Vis. Res.*, vol. 19, no. 10, pp. 1093–1104, 1979.
[46] J.B. Jonas, U. Schneider, and G.O.H. Naumann, "Count and density of human retinal photoreceptors," *Graefe's Arch. Clin. Exp. Ophthalmol.*, vol. 230, pp. 505-510, 1992.
[47] D. E. Irwin, "Visual Memory Within and Across Fixations," *Eye movements and Visual Cognition*, 1992.
[48] W. Li, M.F. Goodchild, and R. Church, "An efficient measure of compactness for two-dimensional shapes and its application in regionalization problems," *International Journal of Geographical Information Science*, vol. 27, 2013.
[49] M. Wopking, "Viewing comfort with stereoscopic pictures," *Journal of the SID*, 1995.
[50] P. Viola , M. Jones, "Rapid object detection using a boosted cascade of simple features," CVPR 2001.
[51] P. F. Felzenszwalb et al., "Object detection with discriminatively trained part based models," PAMI 2010.
[52] M. Everingham et al., "The PASCAL visual object classes challenge - a retrospective," *International Journal of Computer Vision*, 2014.
[53] R. Smith, "An overview of Tesseract OCR engine," *Proc. Ninth Int. Conference on Document Analysis and Recognition (ICDAR)*, 2007.
[54] A. Oliva and A. Torralba, "Modeling the shape of the scene: A holistic presentation of the spatial envelope," *International Journal of Computer Vision*, vol. 42, pp. 145-175, 2001.
[55] C. Shen, M. Song and Q. Zhao, "Learning high-level concepts by training a deep network on eye fixations," Deep Learning and Unsupervised Feature Learning Workshop, USA, December 2012.
[56] L. Breiman and A. Cutler, "Random forest" *Machine Learning* vol. 45, pp. 5-32, 2001.
[57] A. Banitalebi-Dehkordi, E. Nasiopoulos, M. T. Pourazad, and Panos Nasiopoulos, "Benchmark three-dimensional eye-tracking dataset for visual saliency prediction on stereoscopic three-dimensional video," *Journal of Electronic Imaging*, 25 (1), 013008 (January 14, 2016); doi: 10.1117/1.JEI.25.1.013008. Data available at: http://dml.ece.ubc.ca/data/
[58] P. Hanhart and T. Ebrahimi, "EyeC3D: 3D video eye tracking dataset," QoMEX 2014, Singapore.
[59] Y. Fang et al., "An eye-tracking database for stereoscopic video," QoMEX 2014, Singapore.
[60] D. Xu, L. E. Coria, and P. Nasiopoulos, "Guidelines for an Improved Quality of Experience in 3D TV and 3D Mobile Displays," *J. SID*, vol. 20, no. 7, pp. 397–407, July 2012, doi:10.1002/jsid.99.
[61] SensoMotoric Instruments (SMI), "Experiment center  manual," 2010.
[62] A. Borji, D. N. Sihite, and L. Itti, "Quantitative analysis of human-model agreement in visual saliency modeling: A comparative study," *IEEE TIP*, vol. 22, no. 1, pp.55-69, 2013.
[63] Y. Rubner, C. Tomasi, and L. J. Guibas, "The earth movers distance as a metric for image retrieval," *Int. J.  Computer Vision*, 40:2000, 2000.
[64] K. P. Burnham and D. R. Anderson, "Model selection and multi-model inference," *Springer*. (2nd ed), p.51.2002.
[65] T. Judd, F. Durand, and A. Torralba, "A benchmark of computational models of saliency to predict human fixations," Comp. Sci. and Artificial Intelligence Lab. Tech. Report, 2012. http://saliency.mit.edu/
[66] N. Riche et al., "Saliency and Human Fixations: State-of-the-Art and Study of Comparison Metrics," ICCV 2013, pages 1153-1160.
[67] L. Coria, D. Xu, and P. Nasiopoulos, "Automatic stereoscopic 3D video reframing," 3DTV 2012, ETH Zurich, Oct. 15-17, 2012, pp. 1-4.
[68] Y. Park, B. Lee, W. Cheong, and N. Hur, "Stereoscopic 3D visual attention model considering comfortable viewing," IET IPR 2012.
[69] X. Fan, Z. Liu, and G. Sun, "Salient region detection for stereoscopic images," 19[th] International Conference on DSP, 2014, Hong Kong.
[70] R. Ju, L. Ge, W. Geng, T. Ren, and G. Wu, "Depth saliency based on anisotropic center-surround difference," ICIP 2014.
[71] Q. Jiang, F. Duan, and F. Shao, "3D Visual attention for stereoscopic image quality assessment," *Journal of Software*, vol. 9, no. 7, July 2014.
[72] L. Breiman, "Random Forests". Machine Learning. 45 (1): 5–32. doi:10.1023/A:1010933404324, 2001.
[73] A. Banitalebi-Dehkordi, M. T. Pourazad, and P. Nasiopoulos, "A human visual system based 3D video quality metric," 2[nd] International Conference on 3D Imaging, IC3D, Dec. 2012, Belgium.
[74] A. Banitalebi-Dehkordi, M. T. Pourazad, and P. Nasiopoulos, "3D video quality metric for mobile applications," 38[th] International Conference on Acoustic, Speech, and Signal Processing, ICASSP, May 2013, Vancouver, Canada.
[75] A. Banitalebi-Dehkordi, M. T. Pourazad, and P. Nasiopoulos, "3D video quality metric for 3D video compression," 11[th] IEEE IVMSP Workshop: 3D Image/Video Technologies and Applications, June 2013, Seoul, Korea.
[76] A. Banitalebi-Dehkordi, M.T. Pourazad, and Panos Nasiopoulos, "An efficient human visual system based quality metric for 3D video," Springer Journal of Multimedia Tools and Applications, pp. 1-29, Feb. 2015, DOI: 10.1007/s11042-015-2466-z.
[77] W. Chen, J. Fournier, M. Barkowsky, and P. Le Callet, "Quality of Experience Model for 3DTV," SPIE, San Francisco, USA, 2012.
[78] A. Banitalebi-Dehkordi, M. Azimi, M. T. Pourazad, and P. Nasiopoulos, "Compression of high dynamic range video using the HEVC and H. 264/AVC standards," QSHINE 2014 Conference, Greece, Aug. 2014 (invited paper).
[79] M. Azimi, A. Banitalebi-Dehkordi, Y. Dong, M. T. Pourazad, and P. Nasiopoulos, "Evaluating the performance of existing full-reference quality metrics on High Dynamic Range (HDR) Video content," ICMSP 2014: XII International Conference on Multimedia Signal Processing, Nov. 2014, Venice, Italy.
[80] A. Banitalebi-Dehkordi, Y. Dong, M. T. Pourazad, and Panos Nasiopoulos, "A Learning Based Visual Saliency Fusion Model For High Dynamic Range Video (LBVS-HDR)," 23[rd] European Signal Processing Conference, EUSIPCO 2015.
[81] A. Banitalebi-Dehkordi, M. Azimi, M. T. Pourazad, and P. Nasiopoulos, "Visual saliency aided High Dynamic Range (HDR) video quality metrics," 2016 IEEE International Conference on Communications Workshops (ICC).
[82] D. Khaustova et al., "How visual attention is modified by disparities and textures changes?" SPIE HVEI, 2013.
[83] B. Schauerte and R. Stiefelhagen, "Quaternion-based spectral saliency detection for eye fixation prediction," ECCV 2012.
[84] T. Yubing, F.A. Cheikh, F.F.E. Guraya, H. Konik, A. Tremeau, "A spatiotemporal saliency model for video surveillance," Cognitive Computation, vol. 3, pp. 241-263, 2011.
[85] L. Zhang, Zh. Gu, and H. Li, "SDSP: a novel saliency detection method by combining simple priors," ICIP, 2013.
Breiman, Leo (2001). "Random Forests". Machine Learning. 45 (1): 5–32. doi:10.1023/A:1010933404324.